\documentclass[runningheads]{llncs}
\usepackage[T1]{fontenc}
\usepackage{graphicx}
\usepackage[sort]{cite}
\newenvironment{inlinelist}{\begin{enumerate*}[label=\emph{(\roman{*})}]}{\end{enumerate*}}

\newenvironment{inlinelist3}{\begin{enumerate*}[label=\emph{(\arabic{*})}]}{\end{enumerate*}}
\usepackage[inline]{enumitem}
\usepackage{algpseudocode}
\usepackage{adjustbox}
\usepackage{algorithm}
\usepackage[hidelinks]{hyperref}
\makeatletter
\newcommand{\printfnsymbol}[1]{\textsuperscript{\@fnsymbol{#1}}}
\makeatother
\usepackage[symbol]{footmisc}

\begin{document}
\sloppy
\title{``Stop replacing salt with sugar!'': \\ Towards Intuitive Human-Agent Teaching}
\titlerunning{Towards Intuitive Human-Agent Teaching}
\author{Nikolaos Kondylidis\inst{1}\thanks{Co-first authors}\orcidID{0000-0003-4304-564X} \and
Andrea Rafanelli\inst{2}\printfnsymbol{1}\orcidID{0000-0001-8626-2121} \and
Ilaria Tiddi\inst{1}\orcidID{0000-0001-7116-9338} \and
Annette ten Teije\inst{1}\orcidID{0000-0002-9771-8822} \and
Frank van Harmelen\inst{1}\orcidID{0000-0002-7913-0048}}
\authorrunning{Kondylidis et al.}
\institute{Vrije Universiteit Amsterdam,
Amsterdam, The Netherlands \\
\email{\{nikos.kondylidis,i.tiddi,annette.ten.teije,frank.van.harmelen\}@vu.nl}\\ \and
University of Pisa, Pisa, Italy\\
\email{andrea.rafanelli@phd.unipi.it} }
\maketitle             

\vspace{-0.5cm}

\begin{abstract}
Humans quickly learn new concepts from a small number of examples.
Replicating this capacity with Artificial Intelligence (AI) systems has proven to be challenging.
When it comes to learning subjective tasks--where there is an evident scarcity of data--this capacity needs to be recreated.
In this work, we propose an intuitive \textit{human-agent teaching} architecture in which the human can teach an agent how to perform a task by providing demonstrations, i.e., examples.
To have an intuitive interaction, we argue that the agent should be able to learn incrementally from a few single examples.
To allow for this, our objective is to broaden the agent's task understanding using domain knowledge.
Then, using a learning method to enable the agent to learn efficiently from a limited number of examples.
Finally, to optimize how human can select the most representative and
less redundant examples to provide the agent with.
We apply our proposed method to the subjective task of ingredient substitution, where the agent needs to learn how to substitute ingredients in recipes based on human examples. 
We replicate human input using the Recipe1MSubs dataset.
In our experiments, the agent achieves half its task performance after only 100 examples are provided, compared to the complete training set of 50k examples.
We show that by providing examples in strategic order along with a learning method that leverages external symbolic knowledge, the agent can generalize more efficiently.

\end{abstract}

\vspace{-0.5cm}

\vspace{-0.3cm}

\vspace{-0.3cm}
\section{Introduction}
\label{sec:intro}
\vspace{-0.2cm}
One of the most remarkable abilities of the human mind is its capacity to quickly learn new concepts from a small number of examples \cite{Lake2019}.
This learning ability has proven extremely difficult to replicate in Artificial Intelligence (AI) systems.
Indeed, much of the knowledge acquisition happens during training, when models are exposed to large amounts of data.
This is especially true given that most learning processes rely on greedy Machine Learning (ML) approaches, which need many annotated data to effectively learn new concepts~\cite{Adadi21,Alpaydın21,Rafanelli23}.
However, in certain scenarios, there is an evident problem of data scarcity, such as tasks that require personalized or sensitive data.
For instance, while recommendation systems can rely on a consistent amount of user data, there are many domains--such as healthcare, nutrition, education, and so on--where per-user data is not readily available.
Indeed, training an agent for private sensitive tasks, in a way that is tailored to individual preferences and styles--i.e., subjective tasks--becomes challenging.
One promising solution is to allow the agent to be trained by interacting with a human, enabling it to learn based on user behavior.
%
% Under this perspective, a human can serve as a teacher, helping the agent to improve, or accelerate its learning process, or learn new concepts efficiently.

Several studies explore the interaction between humans and agents, \textit{Human-Agent Interaction} (HAI), where the agent needs to learn from the human~\cite{Fails03,CruzI20}. 
%
%For example, in \textit{Interactive Machine Learning} (IML) \cite{Fails03}, humans integrate their knowledge into the learning process to improve or personalize the agents' behavior \cite{CruzI20}.
%
In many cases, the human acts as an optional or supplementary module that can improve the learning process.
Instead, we focus on tasks where human input is the main training signal.
We argue that, in the context of subjective tasks, it is crucial that the agent adjusts to individual user preferences over time. 
In a sense, ``competencies cannot be fixed before deployment and agents must adapt and learn through ongoing interaction with users'' \cite{AkataBRDDEFGHHH20}.
Adapting to user preferences ensures that the agent can dynamically respond to user needs, adjust its behavior, and provide appropriate support, promoting \textit{mutual responsiveness} \cite{Cila22} and, consequently, effective collaboration between the human and the agent.

The study of \cite{KondylidisTT23} presents a framework for efficient human-agent communication by interacting over grounded examples.
Following these guidelines, we propose a \textbf{human-agent teaching} architecture that aims to be intuitive.
Our premise is that, to be intuitive, the human should provide demonstrations of how the task can be performed, and the agent should be able to learn incrementally (adapting), i.e. to exhibit learning outcomes after each demonstration, and quickly (efficiently), i.e. to minimize human input and overall effort.
We reformulate these requirements by envisioning three different components in our architecture, each paired with a specific research question:
\vspace{-0.2cm}

\begin{enumerate} 
\item \textbf{Domain Knowledge}:
\begin{enumerate*}[label=\textbf{(RQ1)}]
\item \label{rq1} \textit{Which external knowledge assists generalization?}
\end{enumerate*}

\item \textbf{Learning Method}: 
\begin{enumerate*}[label=\textbf{(RQ2)}]
\item \label{rq2} \textit{What is an efficient incremental learning method?}
\end{enumerate*}

\item \textbf{Tutoring Policy}: 
\begin{enumerate*}[label=\textbf{(RQ3)}]
\item \label{rq3} \textit{What tutoring policy best selects informative examples for agent learning?}
\end{enumerate*}
\end{enumerate}
\vspace{-0.2cm}

%
% More specifically, by introducing \textbf{domain knowledge}, we aim to broaden the agent’s task understanding beyond its initial training examples data. 
% %
% The idea is to speed up the agent’s learning process by providing it with a larger variety of data. 
% %
% Then, for the \textbf{learning method}, we aim to %leverage on few-shot learning ideas to 
% enable the agent to learn efficiently from a limited number of examples.
% %
% Finally, for the \textbf{tutoring policy}, we aim to %take inspiration from active learning concepts to 
% optimize the way in which the human can select the most representative and less redundant examples.

Through the application of our architecture, we want to build a human-agent teaching scenario where:
\begin{inlinelist3}
    \item the domain knowledge can enable the agent to pick up the task faster \ref{rq1},
    \item the agent exhibits efficient and incremental learning behavior \ref{rq2}, and
    \item the user should provide examples to the agent in a strategic order for faster generalization \ref{rq3}.
\end{inlinelist3}
Our approach is conceptually inline with recent studies on intuitive human-robot tutoring via examples or demonstrations that gain increasing momentum~\cite{MirskyBFHXYIG23,PaciTPB23}, while we also present experimental findings.
Indeed, to apply our architecture, we consider an experimental use case where the agent has to learn how a human substitutes ingredients in recipes, using as few provided examples as possible.
We decided to focus on the subjective task of ingredient substitution, as it may depend on personal preferences, dietary restrictions, etc. \cite{Fatemi2023}.
For this task, the agent can incorporate and utilize generic domain knowledge, like ingredient properties or similarities, but the main training signal must come from its user in terms of examples of how they substitute ingredients, e.g. ``we can substitute butter with oil in this recipe''.
The examples are formed using a shared vocabulary of ingredients and are not conveyed in natural language.

To answer \ref{rq1}, we test different prior generic ingredient knowledge originating either from symbolic knowledge or from pre-trained representations of ML models, as described in Sec. \ref{rq1_ingredient_perception_methods}.
To answer \ref{rq2}, we develop different learning methods (Sec. \ref{sub:learning}) enabling the agent to learn efficiently from a limited number of examples.
Finally, for \ref{rq3}, we propose a tutoring method (Sec. \ref{sub:balanced_tutoring_policy}) %taking inspiration from \textit{active learning} concepts 
to optimize how the user can select the most representative and less redundant examples to provide next.
This tutoring method aims to speed up the agent's learning curve, and it is not meant to replicate how humans select examples, nor it is based on cognitive models. 
Instead, it is meant to provide teaching guidelines to a human user. 
% for the human on how to teach an agent.
% This computationally inspired method is meant to provide informative examples and is not based on any cognitive models, nor is meant to replicate how humans teach, but just to provide informative examples in our use-case.
% 
% \footnote{Note that we do not mean to take inspiration from any cognitive model, or to replicate how humans teach or select examples, but just to provide informative examples in our use-case.}.
%
For our experiments, we use ingredient substitution samples from the \textit{Recipe1MSubs} \cite{Fatemi2023} dataset to synthetically replicate example-based interactions with a human user.

The key contribution of this paper is an architecture for intuitive human-agent teaching.
Moreover, from its application and answering our RQs there are three more contributions:
% scenario for the use-case of ingredient substitution. 
%
% Specifically:
\begin{inlinelist}
    \item We demonstrate how symbolic knowledge can enable the agent to pick up the task faster \ref{rq1};
    \item We suggest a method that exhibits efficient and incremental learning behavior \ref{rq2};
    \item We propose a strategic order the user should provide examples to the agent for faster generalization \ref{rq3}.
\end{inlinelist}

% The contributions of this paper are the following.
% % 
% First, an architecture for intuitive human-agent teaching, capturing how a human user can teach an agent how it wants some tasks to be performed.
% % 
% Second, the application of this framework on the su

% The key contributions of this paper  is an intuitive human-agent teaching architecture and its application the 
% % 

% The architecture 
% where humans teach subjective tasks through examples of desired behavior, which the agent observes and learns from, and the application of this architecture to the use-case of ingredient substitution.
%

This paper is organized as follows.
Sec. \ref{sec:related_work} reports related works and methods that inspire our study.
Sec. \ref{sec:architecture} describes the human-agent teaching architecture and its components.
Sec. \ref{sub:use_case} presents the application of this architecture to the use-case of ingredient substitution and the methods proposed for each RQ.
% present the application of this ac introduces our ingredient substitution use-case.
% % In Sec. \ref{sub:use_case} we present the application of this ac introduces our ingredient substitution use-case.
% %
% Sec. \ref{sub:perception} presents various representations of ingredients.
% %
% Sec. \ref{sub:learning} introduces different learning methods used by the agent to incrementally link appropriate target ingredients to source ingredient representations.
% %
% Sec. \ref{sub:balanced_tutoring_policy} investigates an optimal example sequence for improving the agent's effectiveness.
%
Sec. \ref{sec:experiments} provides experimental results and discusses the main findings.
Finally, Sec. \ref{sec:conclusion} concludes the paper.

\vspace{-0.3cm}
%%%%%%%%%%%%%%%%%%%%%%%%%%%%%%%%%%%%%%%%%%%%%%%%%%%%%%%%%%%%%%%%%%%%%%%%
%%%%%%%%%%%%%%%%%%%%%%%%%%%%%%%%%%%%%%%%%%%%%%%%%%%%%%%%%%%%%%%%%%%%%%%%%
\section{Background and Related Works}
\label{sec:related_work}
\vspace{-0.2cm}
This section provides a review of the literature relevant to our proposed architecture. 
First, we examine the approaches that inspired its design. 
Next, we delve into its key components.
For each key component, we review techniques relevant to that component's implementation.

\vspace{-0.3cm}
\subsection{Architecture}
\vspace{-0.1cm}
\paragraph{Human-Agent Interaction}
has gained significant attention in the past 20 years ~\cite{AkataBRDDEFGHHH20,BarrettRKS17,MohanL14,LiMM20}.
An important area of this field, referred as Learning from Demonstration (LfD)~\cite{Schaal96,HouHEB23}, focuses on the use of human demonstrations to train agents.
The agent learns to mimic a task by observing human examples.
%
%Among these, one relevant approach is the one of inverse reinforcement learning \cite{Russell98}, where the agent tries to extract and learn a reward function from human behavior and use it to derive an imitation policy  ~\cite{ImaniN19, KrishnanGLTMPG19}.
%
Other methods focus more on refining the agent's learning by exploiting human feedback and inputs.
For instance, in Interactive Machine Learning (IML) \cite{Fails03}, the human helps the agent to refine or improve its knowledge through interactions.
In particular, the human observes the data, extracts meaning and insights from them, and then incrementally integrates the gained knowledge into the model (or into the agent) \cite{JiangLC19}.
%
%In interactive reinforcement learning \cite{CruzI20}, human feedback can be used to correct agent's actions or provide different rewards/penalties during the learning process.
%
Our architecture shares similarities with IML and LfD.
It shares with LfD the idea of learning from human examples; however, our approach requires an ongoing and incremental interaction between humans and agents with one example at a time.
Regarding IML, we share the same iterative and incremental interaction goal.
Rather, our agent learns how to complete the task directly from human examples, starting with domain knowledge but lacking task-performing labels, i.e., it does not have access to a set of training data beforehand.
IML, on the other hand, uses both human feedback and training data to help the agent learn.
\vspace{-0.2cm}
\paragraph{Case-Base Reasoning}
(CBR) can offer a powerful tool for human-agent teaching scenarios, enabling agents to draw on past user interactions to inform current decisions.
CBR is a problem-solving approach that involves adapting past solutions to solve new situations~\cite{Riesbeck2013,WatsonM94}.
This approach shares some similarities with our approach.
First, CBR is an approach to incremental and sustained learning, since a new experience is retained each time a problem has been solved \cite{AamodtP94}.
Our architecture also takes into account incremental learning.
Second, CBR solves a new problem by finding similar past ``cases''. 
How the agent learn in our case is similarly driven by looking for similar previous examples that the agent has received.
CBR requires some adaptation mechanisms, as it is often necessary to adapt an old solution to a new situation \cite{Kolodner92}. 
In our case, the agent makes decisions directly based on past examples, without applying any adaptation based on differences between past examples and the current one.
%
%Moreover, CBR does not aggregate past experiences as we do.
\vspace{-0.2cm}
\paragraph{User Cold-Start Recommendation}
% Extending beyond these approaches, our work also connects to challenges faced in recommendation systems.
% %
% A user is using a recommendation system to get suggestions on items they might like (e.g., movies or restaurants).
% %
% Such systems work under the assumption that ``similar users will enjoy similar items''.
%
problem  \cite{cold_start_recommendation} appears when a new user is introduced to the system with very few item ratings and the system must find similar users to recommend items accordingly.
This is a similar problem to our architecture since we aim to learn how a user would perform a subjective task given only a few examples.
Nevertheless, we focus on privacy-sensitive tasks, which restricts us from assuming that we have data from other users for the task.

%%%%%%%%%%%%%%%%%%%%%%%%%%%%%%%%%%%%%%%%%%%%%%%%%%%%%%%%%%%%%%%%%%%%%%%%%
\vspace{-0.2cm}
\subsection{Key Components}
\vspace{-0.1cm}
\paragraph{Knowledge Injection \ref{rq1}}
 is a Neuro-Symbolic (NeSy) technique~\cite{Hitzler22,BekkumBHMT21,BesoldGBBDHKLLPPPZ21} used for enhancing learning systems' performance by incorporating symbolic knowledge.
These approaches are mainly divided into three categories:
\begin{inlinelist}
    \item loss-based techniques ~\cite{DiligentiRG17,XuZFLB18,DiligentiGS17}, that incorporate the logic constraints into the loss function of Deep Learning (DL) models;
    \item structure-based techniques ~\cite{FrancaZG14,GarcezG04}, that build specialized neural structures to incorporate the logic constraints into the neural network's structure; and
    \item embedding-based techniques ~\cite{WuFCABW18,ZhouYJ21}, that generate numeric data from symbolic ones to incorporate additional domain-specific knowledge \cite{ARMCO23}.
\end{inlinelist}
In our architecture, we envision the use of these techniques to integrate domain knowledge into the agent system.
Our goal is to extend the agent's understanding beyond the examples provided by the human, by adding domain knowledge.
The purpose is to give the agent more diverse knowledge to accelerate its learning process.

%%%%%%%%%%%%%%%%%%%%%%%%%%%%%%%%%%%%%%%%%%%%%%%%%%%%%%%%%%%%%%%%%%%%%%%%
\vspace{-0.2cm}
\paragraph{Few-Shot Learning (FSL) \ref{rq2}}
\label{rw:few_shot}
aims to create models that can learn from just a few instances, overcoming the need for a large amount of annotated examples \cite{Parnami22}.
Numerous approaches have been proposed in this area (see \cite{Song0CMS23,TianLLRNT24,Jadon20}).
%
% One straightforward categorization is the one of \cite{Jadon20}, that divides FSL methods into four main groups:
% %
% \begin{inlinelist}
%     \item data augmentation methods,
%     \item metric-based methods,
%     \item model-based methods, and
%     \item optimization-based methods.
% \end{inlinelist}
%
In this work, we mainly refer to \textit{metric-based techniques}  ~\cite{VinyalsBLKW16,SnellSZ17} that leverage distance metrics to measure similarity between samples to classify objects in the embedding space.
One promising method here is the one of prototypical networks \cite{SnellSZ17}, which cluster similar instances in an embedding space. 
For each class, this method computes a prototype by averaging support example embeddings, then classifies queries based on distance to prototypes. 
In our architecture, the agent cannot be trained on how to perform a task before interacting with the human and has to learn from a few examples right from the start.
As we cannot apply prototypical networks as defined in \cite{SnellSZ17}, instead we implement an adapted version that averages representations of instances per class, as a form of training.
%
%%%%%%%%%%%%%%%%%%%%%%%%%%%%%%%%%%%%%%%%%%%%%%%%%%%%%%%%%%%%%%%%%%%%%%%%
\vspace{-0.3cm}
\paragraph{Active Learning (AL) \ref{rq3}}
methods are useful to allow for the selection of the most informative examples for learning.
%
%This field studies how learners can learn from data provided by an oracle.
%
These approaches allow to a shift from a passive learning process to an active one, where the learner can make decisions to improve its understanding of the task \cite{BalcanHV10}.
The key feature is that if the learner has access to some initial training data and many unlabeled data, then it can actively ask the oracle about specific data to label \cite{BalcanF13}.
In our architecture, the increased interactions between human and agent need a better understanding of how human involvement affects the learning process \cite{AmershiCKK14}.
Specifically, the human must select examples for the agent to learn efficiently.
Here, we do not directly make use of AL methods, since we model a situation in which the agent does not have access to unlabeled data.
Consequently, it cannot choose from which example wants to learn next.
Instead, we create a strategy from the human perspective that allows, thanks to understanding of the task, to choose which example to provide next.
We take inspiration from AL ideas to design a strategy that provides examples that are most informative and beneficial for the agent's learning process.

%%%%%%%%%%%%%%%%%%%%%%%%%%%%%%%%%%%%%%%%%%%%%%%%%%%%%%%%%%%%%%%%%%%%%%%%

%%%%%%%%%%%%%%%%%%%%%%%%%%%%%%%%%%%%%%%%%%%%%%%%%%%%%%%%%%%%%%%%%%%%%%%%
%%%%%%%%%%%%%%%%%%%%%%%%%%%%%%%%%%%%%%%%%%%%%%%%%%%%%%%%%%%%%%%%%%%%%%%%
\vspace{-0.3cm}
\section{An architecture for Intuitive Human-Agent Teaching}
\vspace{-0.2cm}
\label{sec:architecture}

Our architecture aims to enable an interaction between a human and an agent, through which the agent learns to perform a subjective task by imitating human behavior.
The human demonstrates a task by providing examples of desired behavior, which the agent observes and learns from.
In this sense, the human acts as a \textbf{teacher}, guiding the agent's learning process through examples.
To enable this interaction, we follow \cite{KondylidisTT23}, which provides guidelines for designing and establishing efficient human-agent task-oriented communication.
Specifically, the framework outlines the cycle of interactions between two agents, where one agent, referred to as \textbf{teacher}, can understand the task at hand and provide examples to the other agent, the \textbf{student}.
Meanwhile, the \textbf{student} needs to comprehend the task and complete it by interacting with the \textbf{teacher}.
The framework describes the following elements:
\begin{inlinelist}
    \item \textit{grounded communication} among the agents, common objects that both can recognize;
    \item \textit{cooperation} among the agents, i.e., none of them can complete the task on their own;
    \item \textit{efficiency}, i.e., time or cognitive load should be minimized.
\end{inlinelist}
We apply these guidelines for our architecture by:
\begin{inlinelist}
    \item defining a shared set of identifiable objects to which the agents may relate, 
    \item outlining an activity that requires cooperation between the two agents to complete,
    \item reducing the number of interactions between the two agents so that the human may give the agent a limited number of specific examples.
\end{inlinelist}
\begin{figure*}[!ht]
\vspace{-0.5cm}
    \centering
    \includegraphics[height=0.4\linewidth, width=0.9\textwidth]{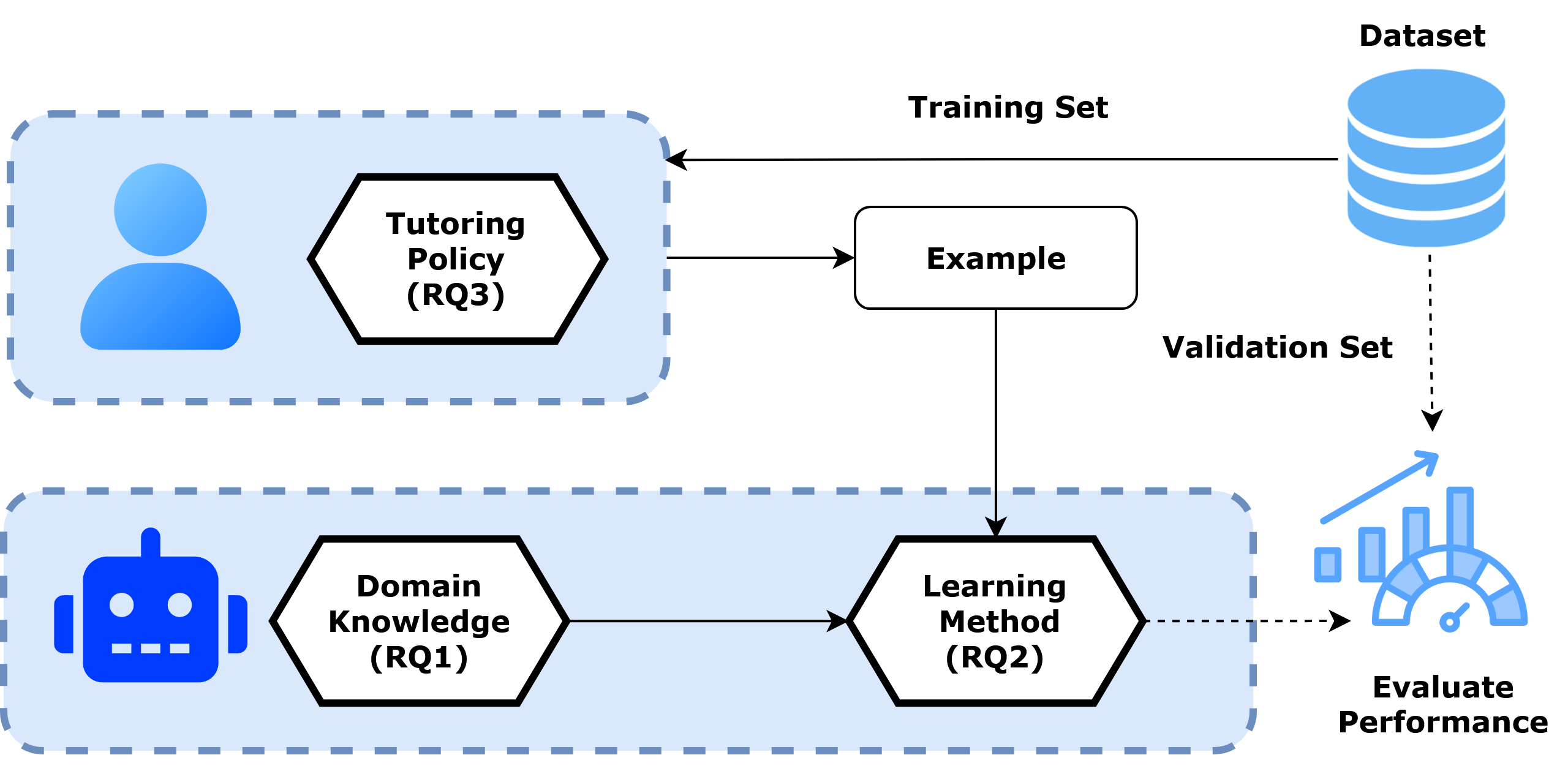}
    \caption{\centering Human-agent teaching architecture.
    Human input (top left) uses the training split of the dataset to provide demonstrative examples to the agent in the order that the \textit{Tutoring Policy} dictates.
    The agent (bottom left) utilizes \textit{Domain Knowledge} to broaden its understanding beyond its initial training examples and data and applies a \textit{Learning Method}, aiming to learn and generalize fast.
    Every few provided training examples (continuous arrows), the agent is evaluated on the complete validation set of the dataset (dashed arrows).} 
    \label{fig:architecture}
    \vspace{-0.45cm}
\end{figure*}

Fig. \ref{fig:architecture} shows our human-agent teaching architecture.
We can observe the various components--\textit{Domain Knowledge}, \textit{Learning Methods}, and \textit{Tutoring Policy}--that make up the architecture, where each component is coupled with a specific research question designed to facilitate its development. 
We see two types of interactions:
\begin{inlinelist3}
    \item \textit{training} over single examples (continuous lines), and
    \item \textit{evaluating} the agent's task ability (dashed lines).
\end{inlinelist3}
During training, the next sample from the training split of the dataset is selected according to the \textit{Tutoring Policy}, simulating the next example that the human would provide to the agent.
Here, different policies help measure how example ordering affects learning and identify strategies for faster generalization.
It is important to note that these policies are designed to give the agent examples strategically and provide guidelines for the human to teach the agent, but are not meant to mimic how humans teach.
Through demonstrative examples, the human interacts with the agent.
The agent should be equipped beforehand with additional \textit{Domain Knowledge}, enabling it to generalize fast from a few examples.
%
%To do so, we make use of knowledge injection for providing it with external knowledge it can use and integrate during the learning
%process.
%
Then, the agent employs \textit{Learning Methods} that allow it to learn incrementally from the examples provided by the human.
Every some interactions, we evaluate the agent’s performance on the validation split. 
Agent’s task performances are evaluated in proportion to the number of examples that it has seen, also measuring its efficiency.
This is an important aspect since this process aims to minimize human interactions (i.e. examples provided).

\vspace{-0.3cm}
\section{The Task of Substituting Ingredients}
\label{sub:use_case}
\vspace{-0.2cm}
To apply our architecture, we select a use-case where the human provides examples of ingredient substitutions to an agent, while the agent efficiently learns how the human substitutes ingredients in recipes using as few examples as possible.
We chose this use-case for the following reasons:
\begin{inlinelist3}
    \item ingredient substitution is inherently a \textit{subjective process}, 
    % \footnote{However, we do not assess the task's subjectivity on a per-user basis. This is due to a lack of data, which we leave to subsequent investigations.}
    that each user may perform differently, not having a single optimal solution that adheres to everyone's preferences, dietary restrictions, health aspects, etc., and 
    % with different users performing it differently, and no single optimal solution 
    % % 
    % We argue that there is no a single optimal solution on how to substitute ingredients due to personal preferences, dietary restrictions, health aspects, etc. 
    %
    \item ingredient substitution is a task characterized by \textit{limited user-specific data}, as there is often insufficient data available about individual users' substitution preferences.
\end{inlinelist3}
%Due to the latter reason, we are unable to assess the proposed methods on a per-user basis, and leave this to subsequent investigations.

We apply the guidelines of \cite{KondylidisTT23} characterizing our architecture to the ingredient substitution use-case:
\begin{inlinelist}
    \item \textit{grounded communication}, we define food ingredients as a shared set of identifiable objects,
    \item \textit{cooperation}, we outline the activity of ingredient substitution, focusing on teaching the agent how to perform the task, as it is unable to learn and evaluate ingredient substitutions alone, and
    \item \textit{efficiency}, we want to reduce the interactions between the two agents (human and agent).
    Thus, the human gives the agent a limited number of ingredient substitution examples.
\end{inlinelist}

\vspace{-0.5cm}

%%%%%%%%%%%%%%%%%%%%%%%%%%%%%%%%%%%%%%%%%%%%%%%%%%%%%%%%%%%%%%%%%%%%%%%%%%%

\begin{figure*}[!ht]
\vspace{-0.3cm}
    \centering
    \includegraphics[width=1\textwidth]{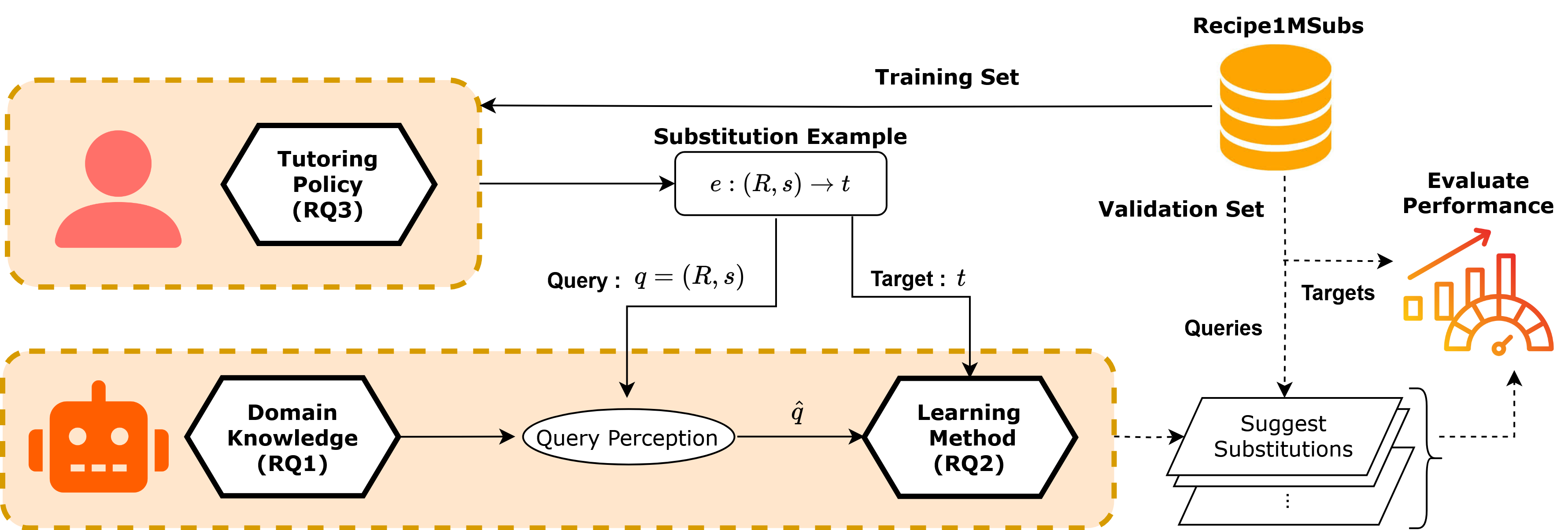}
    \caption{\centering Human-agent interaction for the ingredient substitution task.
    Human input (top left) leverages the Recipe1MSubs dataset's training set to provide substitution examples according to the order determined by the \textit{Tutoring Policy}.
    The agent (bottom left) utilizes \textit{Domain Knowledge} to perceive the substitution example and employs a \textit{Learning Method} to learn incrementally from few examples.
    After few substitution examples provided by the human (continuous arrows), the agent is evaluated on the complete validation set of the dataset (dashed arrows). \vspace{-0.6cm}}
    \label{fig:use_case}
    % \vspace{-0.3cm}
\end{figure*}
\vspace{-0.2cm}
\paragraph{Interaction Design}
\sloppy
In Fig. \ref{fig:use_case}, we show the human-agent teaching architecture applied to the use-case of ingredient substitution.
We use a task-related dataset to mimic human input, i.e., Recipe1MSubs \cite{Fatemi2023}, to train and evaluate an agent without requiring human participants.
The dataset contains $70,520$ ingredient substitutions ($49,044$, $10,729$, and $10,747$ ingredient substitution samples for training, validation, and testing splits, respectively) for $44$k recipes of the Recipe1M \cite{MarinBOHSAWT21} dataset.
Each substitution was crawled from the user-provided comments section of the website where the corresponding recipe was uploaded.
We remove examples where the recipe consists of only the source ingredient, i.e. the ingredient that needs to be substituted, affecting only $29$, $3$, and $1$ samples on training, validation, and test splits, respectively.

Each example is described as $e= (R, s) \rightarrow t$, where $R$ is the set of recipe ingredients, $s$ is the ingredient of the recipe we want to substitute, i.e. the source ingredient, and $t$ is an ingredient that can substitute the source ingredient in this recipe, i.e. the target ingredient.
The recipe ingredients $R$, together with the source ingredient $s$, form the substitution query $q$.
The agent utilizes \textit{Domain Knowledge} and represents the query $q$ in a mathematical space ($\hat{q}$) (Sec. \ref{sub:perception}) that allows generalizing across similar substitution queries \ref{rq1}.
Then it uses a \textit{Learning Method} (Sec. \ref{sub:learning}) that allows it to learn incrementally from single examples, and reach broad ingredient substitution replacement abilities based on limited interactions with the user \ref{rq2}.
The \textit{Tutoring Policy} (Sec. \ref{sub:balanced_tutoring_policy}) is developed in order to strategically select examples that maximize the transfer of ingredient substitution knowledge to the agent within each interaction \ref{rq3}.
%%%%%%%%%%%%%%%%%%%%%%%%%%%%%%%%%%%%%%%%%%%%%%%%%%%%%%%%%%%%%%%%%%%%%%%%%%%%%%%
%%%%%%%%%%%%%%%%%%%%%%%%%%%%%%%%%%%%%%%%%%%%%%%%%%%%%%%%%%%%%%%%%%%%%%%%%%%%%%%
\vspace{-0.3cm}
 \subsection{\ref{rq1} Domain Knowledge} 
 \vspace{-0.1cm}
 \label{sub:perception}

The agent's ingredient and recipe perception, i.e., how it represents ingredients and recipes, can heavily affect its learning efficiency and generalization behavior.
We put forward different ingredient representation methods in Sec. \ref{rq1_ingredient_perception_methods}, allowing us to answer \ref{rq1}.
Then, in Sec. \ref{rq1_recipe_perception_methods}, we show how ingredient representations are aggregated to represent a substitution query.
%%%%%%%%%%%%%%%%%%%%%%%%%%%%%%%%%%%%%%%%%%%%%%%%%%%%%%%%%%%%%%%%%%%%%%%%%%%%%%%
\vspace{-0.3cm}

\subsubsection{Ingredient Representation}
\label{rq1_ingredient_perception_methods}

\paragraph{Orthogonal Ingredient Representation (\textbf{1-hot}):}
\vspace{-0.3cm}

The simplest way to represent the ingredients is to suggest that each ingredient is unique, not related to other ingredients in any way, and equally different from all other ingredients.
This representation method is usually referred to as 1-hot encoding.
The dimensionality of the representation space is equal to the number of ingredients, i.e., 6,632.
This dimensionality is provided by mapping, following \cite{Fatemi2023}, all ingredient names in the dataset to the ingredient vocabulary of FlavorGraph \cite{Park2021} of size 6,632.
%%%%%%%%%%%%%%%%%%%%%%%%%%%%%%%%%%%%%%%%%%%%%%%%%%%%%%%%%%%%%%%%%%%%%%%%%%%%%%%
\vspace{-0.2cm}
\paragraph{Injecting Symbolic Knowledge (\textbf{1-hot \& FoodOn}):}

We incorporate symbolic knowledge by expanding the 1-hot representation with categorical details of ingredients obtained from the FoodOn knowledge graph \cite{foodon2018}.
As an example, one ingredient category is ``{\small \texttt{yellow bean pod}}'', which is a subclass of ``{\small\texttt{plant food product}}''.
FoodOn defines hierarchical concepts related to food, covering the entire journey from farm to table.
Following \cite{ShiraiK22}, we link our ingredients to FoodOn using tf-idf \cite{tf_idf} based on ingredient names and the {\small\texttt{rdfs:label}} property, keeping retaining matches with similarity above 0.6.
We retrieve up to 5 hops of superclasses via the {\small\texttt{rdfs:subClassOf}} property, treating them as more generic ingredient properties.
All properties that describe only one ingredient are ignored as they act as (redundant) 1-hot encodings.
Each property retrieved describes the linked ingredient according to $expressiveness\ weight = 2^{-(\#hops + 1)}$, where $\#hops$ is the number of hops required to retrieve the property.
The initial lexical match counts as the first hop.
The dimensionality of the 1-hot encodings is extended by the 3,463 FoodOn (super)classes utilized for a total dimensionality of 10,116. 
For a given ingredient, the values of the new 3,463 representation dimensions are equal to the expressiveness weight between that ingredient and each FoodOn class, or zero for non-related classes (see the left of Fig. \ref{fig:full_stack_example}).

%%%%%%%%%%%%%%%%%%%%%%%%%%%%%%%%%%%%%%%%%%%%%%%%%%%%%%%%%
\vspace{-0.2cm}
\paragraph{Ingredient Chemistry \& Co-occurrence Embeddings (\textbf{FlavorGraph}):}
FlavorGraph \cite{Park2021} contains co-occurrence and molecular information for 6,653 ingredients represented in a graph format.
Ingredient co-occurrence is calculated on the Recipe1M dataset \cite{MarinBOHSAWT21}, while links with chemical compounds like ``{\small\texttt{Nonanoic Acid}}'' are retrieved from FlavorDB \cite{flavorDB} and HyperFoods \cite{hyperfoods}.
The metapath2vec \cite{DongCS17} algorithm is adapted to train ingredient embeddings on this graph.
This algorithm creates metapaths that propagate chemical information from compound nodes to all ingredient nodes.
A skip-gram model \cite{MikolovSCCD13} is applied to the metapaths to generate node representations.
The adaptation they propose is to introduce a \textit{chemical structure prediction} (CSP) layer on top of the skim-gram model.
We use the produced embeddings that reflect ingredient similarity according to both chemical information and recipe co-occurrence to represent ingredients.
The representation dimensionality is 300.

%%%%%%%%%%%%%%%%%%%%%%%%%%%%%%%%%%%%%%%%%%%%%%%%%%%%%%%%%
\vspace{-0.2cm}
\paragraph{Recipe Instructions Contextualized Embeddings (\textbf{FoodBert}):}
In \cite{PellegriniOWG21}, the authors fine-tune a BERT model \cite{DevlinCLT19} on recipe instructions from Recipe1M \cite{MarinBOHSAWT21} to generate embeddings for food ingredients and recipes.
They expand the model's vocabulary to include ingredients from their dataset, ensuring that each ingredient is represented as a single token.
100 instruction sentences are randomly sampled for each token that represents an ingredient.
The 100 contextualized representations of the token are averaged, resulting in a vector that represents the corresponding ingredient.
We repeat this process for the FlavorGraph ingredient vocabulary, by finetuning the {\small\texttt{bert-base-cased model}} from \textit{Hugging Face} \footnote{\small \url{https://huggingface.co/google-bert/bert-base-cased}} on the Recipe1M dataset.
The aggregated representations of the tokens that represent the 6,632 FlavorGraph ingredients are used directly for ingredient representation.
The representation dimensionality is 768.

% \newpage
%%%%%%%%%%%%%%%%%%%%%%%%%%%%%%%%%%%%%%%%%%%%%%%%%%%%%%%%%
\vspace{-0.3cm}
\subsubsection{Query Perception ($\hat{q}$)}
\label{rq1_recipe_perception_methods}

\begin{figure}[!ht]
\vspace{-0.4cm}
    \centering
    \includegraphics[width=\textwidth, height=0.3\textwidth]{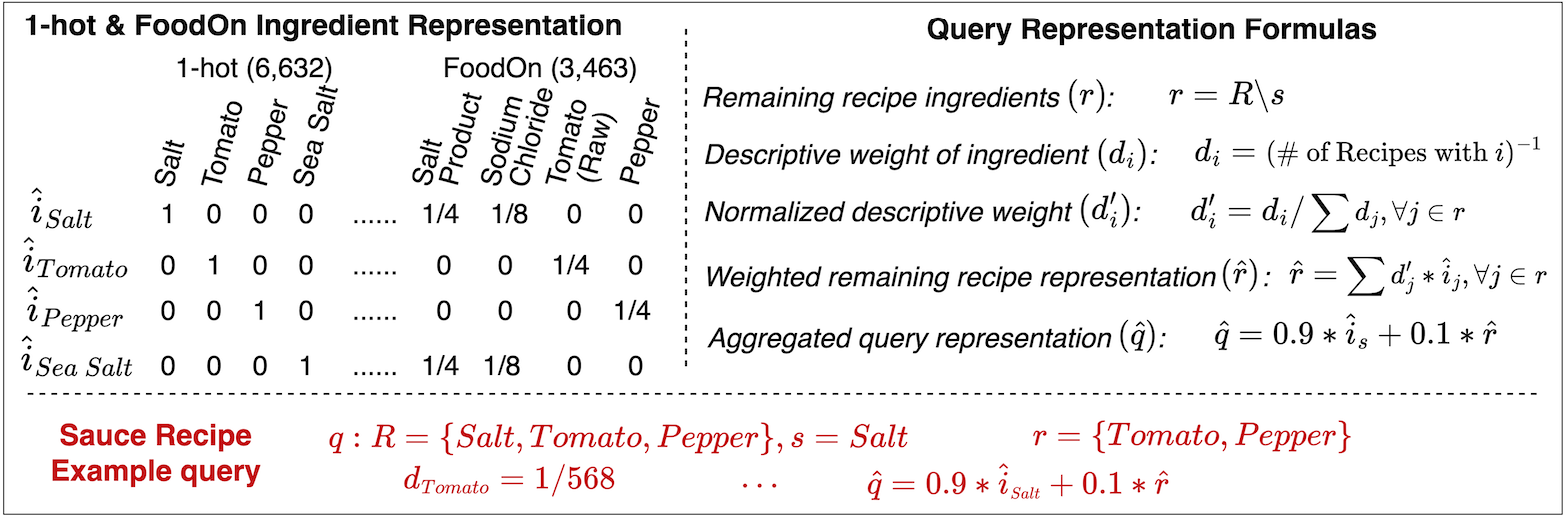}
    \caption{\centering An example of ``1-hot \& FoodOn'' ingredient representation (Left). FoodOn brings Salt and Sea Salt closer compared to only 1-hot representations. Representing a substitution query (replacing Salt, in a sauce recipe with Salt, Tomato, and Pepper), based on ingredient representations (Right). \vspace{-0.6cm}
    }
    \label{fig:full_stack_example}
\end{figure}

\paragraph{}
\vspace{-0.2cm}
Queries are represented as an aggregation of ingredient representations (see right of Fig. \ref{fig:full_stack_example}), regardless of how ingredients are represented.
Let $\hat{i}$ be the vector representation of an ingredient.
We believe that the source ingredient ($s$), i.e., the one we want to replace, has a great impact on shaping the query representation compared to the remaining ingredients of the recipe, i.e., $r = (R\backslash s)$.
Therefore, we decided to give it more weight.
Accordingly, the query representation $\hat{q}$ is a weighted average of the representation of the source ingredient $\hat{s}$ and the representation of the remaining recipe ingredients $\hat{r}$, with weights of 90\% and 10\% accordingly, i.e., $\hat{q} = (0.9*\hat{s} + 0.1*\hat{r})$.
Note that we include $r$ in the query representation to compensate for the lack of dish information in our dataset.
We argue that proper substitutions for an ingredient are influenced by the recipe in which it appears. , i.e., if $s$ is ``oil'' in a cake recipe, it can be substituted with ``butter''. However, in a pizza recipe, ``butter'' would not be an appropriate substitute. 
At the same time, we assume that common ingredients are less descriptive for a recipe compared to rare ones.
Specifically, inspired by \textit{tf-idf} \cite{tf_idf}, we define the \textit{descriptive weight} $d_i$ of an ingredient $i$ as: $d_i=n_i^{-1}$, where $n_i$ is the number of recipes that use ingredient $i$. We used the recipes of the training set of the Recipe1MSubs dataset, although any set of recipes can be used for this calculation.
Then, for a given recipe $R$, we normalize the descriptive weight of its remaining ingredients $r$ to sum to $1$: $d^\prime_i=d_i/(\sum_{j \in r}\, d_j)$, where $j$ iterates over all the remaining ingredients in the recipe $R$.
The representation of the remaining ingredients is:  $\hat{r} = \sum_{j \in r} \hat{i} \cdot d^\prime_i$, where $\hat{i}$ is the representation of ingredient $i$, and the sum is over all the remaining ingredients $j \in r$.
Fig. \ref{fig:full_stack_example} illustrates an example of ingredient representation (left) and how the query representation ($\hat{q}$) is calculated (right). 

%%%%%%%%%%%%%%%%%%%%%%%%%%%%%%%%%%%%%%%%%%%%%%%%%%%%%%%%%%%%%%%%%%%%%%%%
%%%%%%%%%%%%%%%%%%%%%%%%%%%%%%%%%%%%%%%%%%%%%%%%%%%%%%%%%%%%%%%%%%%%%%%%
\vspace{-0.4cm}
\subsection{\ref{rq2} Learning Method}
\label{sub:learning}
%\vspace{-0.2cm}

The presented learning methods can learn incrementally from individual examples.
Their function is to relate query representations to suitable target ingredients.
During inference, they provide a score for each candidate ingredient, allowing the agent to rank their substitution recommendations.
The first one, i.e., \textit{Baseline}, only uses source ingredients to represent the query, $\hat{q}=\hat{s}$.
The other two methods represent the query as $\hat{q} = (0.9*\hat{s} + 0.1*\hat{r})$ (as described in Sec. \ref{rq1_recipe_perception_methods}), for any provided ingredient representation method (see Sec. \ref{rq1_ingredient_perception_methods}).
Moreover, these two approaches prioritize all ingredients that have been observed as target ingredients during inference, following the assumption that any observed substitute is more likely to be a correct substitute than other ingredients.
%%%%%%%%%%%%%%%%%%%%%%%%%%%%%%%%%%%%%%%%%%%%%%%%%%%%%%%%%%%%%%%%%%%%%%%%
\vspace{-0.2cm}
\paragraph{Source-Target Substitution Frequency (\textbf{\textit{Baseline}}):}
The \textit{Baseline} learning method is the frequency-based method ``LT+Freq'' presented in \cite{Fatemi2023}.
This method counts the frequency each target ingredient $t$ replaces a source ingredient $s$ in the provided examples.
During inference, the method retrieves substitution frequencies for the provided source ingredient and ranks candidate target ingredients from the most to the least frequently observed substitute.
Notably, this approach only takes into account the source ingredient, without considering ingredient properties or recipe context.
We chose this method as a baseline, as it is the best-performing method that can be trained incrementally on individual examples, adhering to the requirements of our approach.

%%%%%%%%%%%%%%%%%%%%%%%%%%%%%%%%%%%%%%%%%%%%%%%%%%%%%%%%%%%%%%%%%%%%%%%%
\vspace{-0.2cm}
\paragraph{Prototypical Networks with pre-trained representation (\textbf{P. Networks}):}

Prototypical networks is a well-established few-shot learning method \cite{SnellSZ17}.
In our case, we do not have a trained model that we use to transfer knowledge or to fine-tune for new classes.
Instead, the agent has to learn the task from scratch from as few examples as possible.
For this reason, we do not directly use prototypical networks, but we do take inspiration from them.
Specifically, we treat each candidate ingredient as a class.
Then, all query representations provided to the agent--in which a target ingredient has been observed as a substitute--compose the cluster (or prototype) of that ingredient.
In practice, target ingredient representations, i.e., the prototype vectors, are equal to the average of their observed query representations, similar to calculating the centroid of a cluster.
By comparing their representations, this method ranks candidate target ingredients during inference based on how similar their prototypes are with the given query representation.
The target ingredient with the most (least) similar representation is ranked at the top (bottom).
\emph{This method leverages the idea that similar substitution queries can use the same substitutes.}

%%%%%%%%%%%%%%%%%%%%%%%%%%%%%%%%%%%%%%%%%%%%%%%%%%%%%%%%%%%%%%%%%%%%%%%%
\vspace{-0.2cm}
\paragraph{\textit{Accumulative} Target Representations (\textbf{\textit{Accumulative}}):}
This method is similar to the one described in the previous paragraph, i.e., Prototypical Networks, but with two differences that aim to give it an advantage.
Averaging over multiple vector representations can lead to loss of useful information, i.e., \textit{semantic dilution}.
To avoid this, we set the representation of target ingredient $\hat{t}$ to be the \textit{addition} of its observed query representations (as compared to the mean used by P. Networks), i.e., $\hat{t} = \hat{t} + \hat{q}$.
In addition, we assume that ingredient substitution is an habitual task and accordingly, we want to promote popular (favourite) target ingredients.
To do this, we estimate the substitution score of a candidate target ingredient $t$ by calculating the inner product between its representation $\hat{t}$ and that of the query $\hat{q}$, i.e., $score = \hat{q} \cdot \hat{t}$.
Since the target ingredient representations are the accumulation of observed query representations, more popular target ingredients will be represented by vectors with larger magnitudes and are expected to have larger inner products with a given query representation.
The value of the score indicates similarity, and the target ingredient with the highest (lowest) score is ranked at the top (bottom).

%%%%%%%%%%%%%%%%%%%%%%%%%%%%%%%%%%%%%%%%%%%%%%%%%%%%%%%%%%%%%%%%%%%%%%%%
%%%%%%%%%%%%%%%%%%%%%%%%%%%%%%%%%%%%%%%%%%%%%%%%%%%%%%%%%%%%%%%%%%%%%%%%
\vspace{-0.2cm}
\subsection{\ref{rq3} Strategic Tutoring Policy}
\label{sub:balanced_tutoring_policy}
%\vspace{-0.2cm}
Since the agent is learning incrementally, the order in which the examples are provided largely affects its learning curve.
For this reason, we investigate what would be a beneficial order to teach the agent and enhance its efficient learning.
The simplest tutoring policy is to provide the examples in \textit{Random} order.
%%%%%%%%%%%%%%%%%%%%%%%%%%%%%%%%%%%%%%%%%%%%%%%%%%%%%%%%%%%%%%%%%%%%%%%%%
\vspace{-0.2cm}
\paragraph{Balancing Exploration and Exploitation (\textbf{Balanced}):}

This tutoring policy aims to balance between exploring new substitution scenarios and ensuring that the most common substitutions remain a dominant pattern.
Considering that the method of counting source-target ingredient frequencies exhibits high task performance, i.e.,``LT+Freq'' in \cite{Fatemi2023}, we decide to balance exploitation and exploration as represented by this signal.
Accordingly, we divide all examples into \textit{buckets} based on their source and target ingredient, i.e., the ``complete buckets''.
Then, we start a nested loop until all examples are used. 
In the first loop, for every complete bucket that has remaining examples, we build a ``reduced bucket''.
The size of the reduced bucket is equal to $floor [log_2(|B|)] + 1$, where $|B|$ is the current size of its corresponding complete bucket, i.e., the number of unused examples with the corresponding source and target ingredients.
In the second nested loop, we pick the largest reduced bucket and randomly select one example from it. 
We then update sizes of both the reduced and complete buckets. 
Once all reduced buckets are empty, we return to the first loop to create new reduced buckets and continue the process.

%%%%%%%%%%%%%%%%%%%%%%%%%%%%%%%%%%%%%%%%%%%%%%%%%%%%%%%%%%%%%%%%%%%%%%%%
%%%%%%%%%%%%%%%%%%%%%%%%%%%%%%%%%%%%%%%%%%%%%%%%%%%%%%%%%%%%%%%%%%%%%%%%
\vspace{-0.4cm}
\section{Experimental Results}
\label{sec:experiments}
\vspace{-0.2cm}
In this section, we describe our experimental setup and the results.
All experimental results are aggregations of four executions.
The agent's task performance is reflected using information retrieval metrics, i.e., hit@k, for $k \in \{1,10\}$ and Mean Reciprocal Rank (MRR).
Hit@$k$ is the ratio of generated substitution suggestions where the target ingredient was found within the top $k$ recommended substitutions.
MRR is the harmonic mean of the retrieved ranks of the target ingredients over all samples of the test set.
How efficiently the agent is picking up the task is measured in terms of the number of examples provided by the human up to that point.

In \cite{Fatemi2023}, when the dataset contains multiple correct substitutions for a substitution query, the highest ranked among them is considered to be the target ingredient.
In contrast, we evaluate each substitution in the dataset separately, allowing an agent to record higher task performance only when all possible ingredient substitutions are highly ranked.
This promotes agents that capture the broad spectrum of possible candidate substitutions in the dataset.
Our code and results are available online\footnote{\small \url{https://github.com/human-agent-teaming/efficient-ingredient-substitution}}.
\vspace{-0.2cm}
\subsection{Domain Knowledge \ref{rq1} \& Learning Methods \ref{rq2}}
%\vspace{-0.1cm}

Table \ref{tab:query_representation_ablation_study} illustrates the performance of different combinations of learning methods and query representation across different metrics.
Notably, \textit{P. Networks} performs poorly with all the different query representations.
Indeed, while showing overall better performances for all the metrics and all the different representations w.r.t. the \textit{Baseline} with $100$ examples provided, as training progresses the performances drastically drop compared to the \textit{Baseline}. 
For example, in the Hit@1 metric, \textit{P.Networks $|$ 1-hot \& FoodOn} starts competitively but then fails to improve or even decline with more examples provided, while the \textit{Baseline} steadily improves from $3.07\%$ to $18.93\%$. 
\textit{We believe that the prototypical representations, i.e., the mean of all observed query representations for each target ingredient, are suffering from semantic dilution when averaging many vectors.}

\vspace{-0.1cm}

\begin{table*}[!ht]
\LARGE
\caption{\centering Performances of different learning methods and ingredient representations. Examples are provided in \textit{Random} order.}
\label{tab:query_representation_ablation_study}
\begin{adjustbox}{max width=\textwidth}
\begin{tabular}{l|ccc|ccc|ccc}
               & \multicolumn{3}{c|}{100 Examples}                                                       & \multicolumn{3}{c|}{10k Examples}                                                       & \multicolumn{3}{c}{All Examples (49k)}                                                 \\ \hline
        \textbf{Learning  $|$ Ingr. Repr.}             & \multicolumn{1}{c}{\textbf{Hit@1\%}} & \multicolumn{1}{c}{\textbf{Hit@10\%}} & \multicolumn{1}{c|}{\textbf{MRR\%}} & \multicolumn{1}{c}{\textbf{Hit@1\%}} & \multicolumn{1}{c}{\textbf{Hit@10\%}} & \multicolumn{1}{c|}{\textbf{MRR\%}} & \multicolumn{1}{c}{\textbf{Hit@1\%}} & \multicolumn{1}{c}{\textbf{Hit@10\%}} & \multicolumn{1}{c}{\textbf{MRR\%}} \\ \hline
Baseline   &   $3.07$                    &        $4.66$                      &       $3.73$                    &           $16.12$                  &              $40.08$                &            $23.93$               &         $18.93$                   &         $50.25$  &           $29.09$     \\

P.Networks $|$ FlavorGraph    &   $2.75$                    &        $7.60$                      &       $4.75$                    &           $1.79$                  &              $19.15$                &            $7.17$               &         $1.54$                   &         $16.61$  &           $6.24$     \\

P.Networks $|$ FoodBert   &   $3.17$                   &        $9.87$                      &       $5.54$                    &           $1.94$                  &               $20.74$               &            $7.70$               &         $1.35$                  &          $17.55$  &           $6.65$     \\

P.Networks $|$ 1-hot    &   $2.52$                    &        $6.68$                      &       $4.30$                    &           $2.83$                  &             $26.81$                 &           $9.79$                &         $2.50$                    &       $25.88$   &          $9.44$    \\

P.Networks $|$ 1-hot \& FoodOn    &   $3.33$                    &        $10.31$                      &       $5.93$                    &           $3.09$                  &             $29.44$                 &           $10.88$                &         $2.72$                    &       $27.23$   &          $10.02$    \\

Accumulative $|$ FlavorGraph    &   $3.66$                    &        $9.29$                      &       $5.84$                    &           $6.91$                  &             $21.88$                 &           $12.01$                &         $7.19$                    &       $21.80$   &          $12.27$    \\

Accumulative $|$ FoodBert    &   $1.97$                    &        $6.53$                      &       $3.73$                    &           $3.08$                  &             $13.38$                 &           $6.47$                &         $3.08$                    &       $13.54$   &          $6.49$    \\
Accumulative $|$ 1-hot    &   $3.07$                    &        $6.72$                      &       $4.56$                    &           $16.32$                  &             $41.80$                 &           $24.87$                &         $\textbf{18.99}$                    &       $51.26$   &          $29.57$    \\
Accumulative $|$ 1-hot \& FoodOn   &   $\textbf{3.82}$                    &        $\textbf{10.49}$                      &       $\textbf{6.25}$                    &           $\textbf{16.90}$                  &             $\textbf{45.52}$                 &           $\textbf{26.43}$                &         $18.85$                    &       $\textbf{52.10}$   &          $\textbf{29.59}$    \\
\end{tabular}
\end{adjustbox}
\vspace{-0.3cm}
\end{table*}

On the other hand, the \textit{Accumulative $|$ 1-hot \& FoodOn}, shows improvement w.r.t. the \textit{Baseline} approach.
This is particularly evident in the first $10$k steps, where for Hit@10 the \textit{Accumulative} approach reaches $41.80\%$ without the inclusion of FoodOn and $45.52\%$ with its inclusion, while the \textit{Baseline} achieves $40.08\%$.
The same behavior is shown for the MRR where it achieves $26.43\%$ with 1-hot \& FoodOn, and $24.87\%$ without, when the \textit{Baseline} reaches $23.93\%$.
However, it is important to note that when this method uses pre-trained embeddings (i.e., FoodBert and Flavorgraph), the performances significantly decrease, indicating that \textit{either the transfer knowledge is not good enough for our task, or there is some limitation on its ability to leverage it.}

Looking at Table \ref{tab:query_representation_ablation_study}, we observe that both the \textit{Accumulative} method with either the 1-hot or the 1-hot \& FoodOn representations, and the \textit{Baseline} learning methods, exhibit behavior of continual learning, aligning with our goal of having an agent capable of incremental learning.
Accordingly, the remaining of our experiments will showcase only the best-performing combination: i.e., \textit{Accumulative $|$ 1-hot \& FoodOn}.
\vspace{-0.2cm}
\subsection{Providing the Examples in Strategic Order \ref{rq3}}
%\vspace{-0.2cm}
\paragraph{Learning from Few Examples.}

\begin{figure*}[h]
\vspace{-0.4cm}
\minipage{0.5\textwidth}
  \includegraphics[width=\textwidth, height=0.7\textwidth]{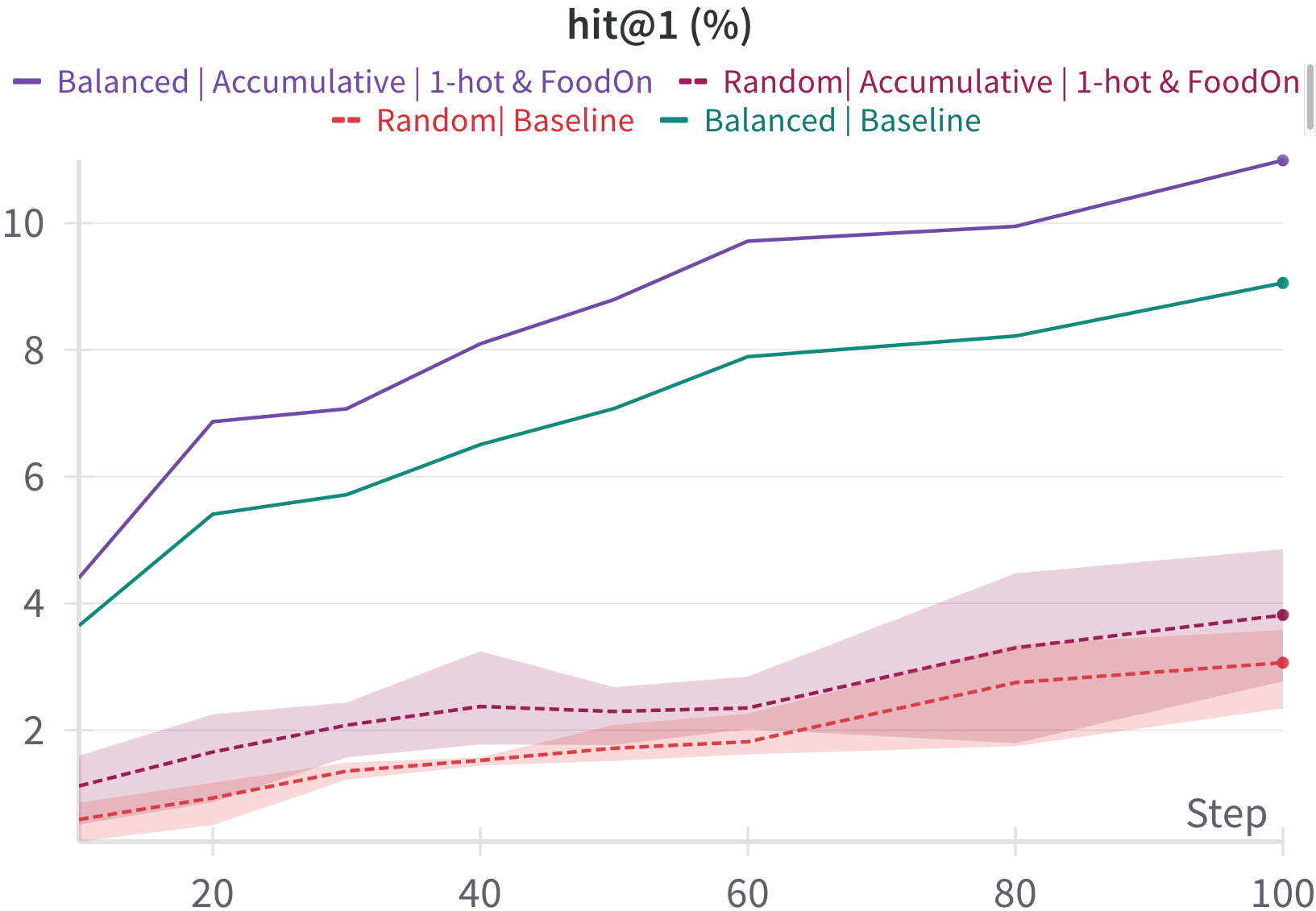}
\endminipage\hfill
\minipage{0.5\textwidth}
  \includegraphics[width=\textwidth, height=0.7\textwidth]{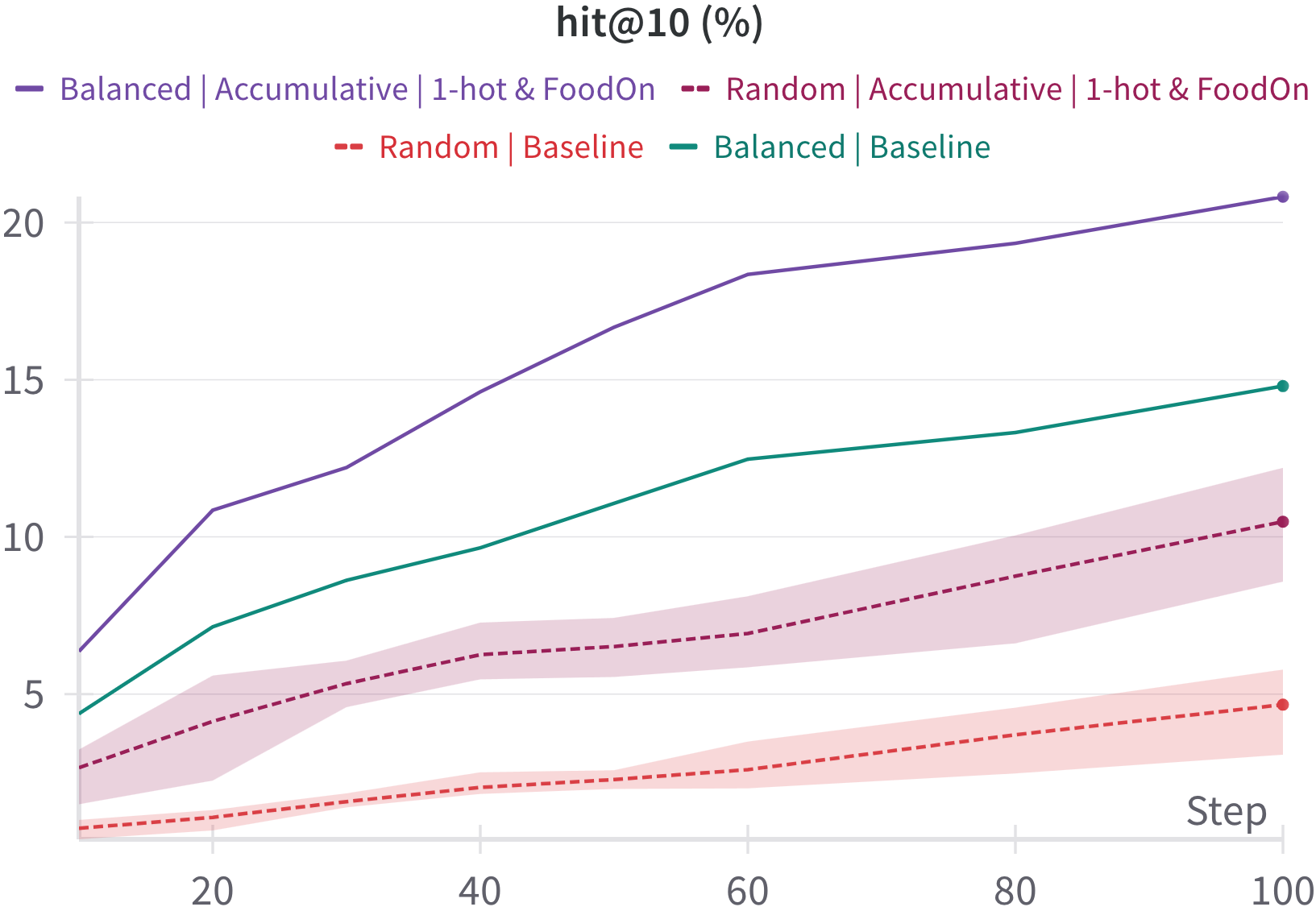}
\endminipage\hfill
\caption{\centering Performance of the \textit{Accumulative $|$ 1-hot \& FoodOn} and the \textit{Baseline} while providing only 100 examples, either in \textit{Random} order, or using the \textit{Balanced} tutoring policy (Sec. \ref{sub:balanced_tutoring_policy}).} 
\label{fig:al_performance_few}
\vspace{-0.4cm}
\end{figure*}

Fig. \ref{fig:al_performance_few} presents the performance of the best-performing method, i.e., \textit{Accumulative $|$ 1-hot \& FoodOn}, and \textit{Baseline} when provided with $100$ \textit{Random} examples or by using the \textit{Balanced} tutoring policy (Sec. \ref{sub:balanced_tutoring_policy}).
When the agent is learning from 100 examples the \textit{Balanced} tutoring policy enables it to double/triple learning performances.
This is also visible by looking at the first column of Table \ref{tab:tutoring-method}.
In fact, it can be seen that when the examples are provided in \textit{Random} order, for both learning methods, the agent performs worst in all metrics w.r.t. when the examples are provided with the \textit{Balanced} tutoring policy.
%examples are provided, the agent performs 3.06, 4.66, and 3,73, when evaluating with Hit@1, Hit@10, and MRR respectively.
%
%When provided with the Balanced tutoring policy, the agent performs 9.05, 14.80, and 11.18, for the same metrics.
%
%For example, when the agent uses \textit{Accumulative $|$ 1-hot \& FoodOn}, the Balanced tutoring policy allows it to achieve from double to almost triple the performance, depending on the evaluation metric.
%
%Hit@1 increases (by 288\%) from 3.82 to 10.99, Hit@10 doubles, reaching 20.82 instead of 10.49, and MRR (is enlarged to 234\%), goes from 6.25 to 14.59.

\begin{table*}[!ht]
\vspace{-0.4cm}
\LARGE
\caption{\centering Performance of the \textit{Accumulative $|$ 1-hot \& FoodOn} and the \textit{Baseline} while providing the entire amount of examples, either in \textit{Random} order or using the \textit{Balanced} tutoring policy (Sec. \ref{sub:balanced_tutoring_policy}).}
\label{tab:tutoring-method}
\begin{adjustbox}{max width=\textwidth}
\begin{tabular}{l|ccc|ccc|ccc}
               & \multicolumn{3}{c|}{100 Examples}                                                       & \multicolumn{3}{c|}{10k Examples}                                                       & \multicolumn{3}{c}{All Examples (49k)}                                                 \\ \hline
        \textbf{Tutoring $|$ Learning  $|$ Ingr. Repr.}             & \multicolumn{1}{c}{\textbf{Hit@1\%}} & \multicolumn{1}{c}{\textbf{Hit@10\%}} & \multicolumn{1}{c|}{\textbf{MRR\%}} & \multicolumn{1}{c}{\textbf{Hit@1\%}} & \multicolumn{1}{c}{\textbf{Hit@10\%}} & \multicolumn{1}{c|}{\textbf{MRR\%}} & \multicolumn{1}{c}{\textbf{Hit@1\%}} & \multicolumn{1}{c}{\textbf{Hit@10\%}} & \multicolumn{1}{c}{\textbf{MRR\%}} \\ \hline
Random $|$ Baseline    &   $3.07$                    &        $4.66$                      &       $3.73$                    &           $16.12$                  &              $40.08$                &            $23.93$               &         $\textbf{18.93}$                   &         $50.25$  &           $29.09$     \\

Balanced $|$ Baseline    &   $9.05$                    &        $14.80$                      &       $11.18$                    &           $17.93$                  &              $46.28$                &            $27.23$               &         $\textbf{18.93}$                   &         $50.25$  &           $29.09$     \\

Random $|$ Accumulative $|$ 1-hot \& FoodOn   &   $3.82$                   &        $10.49$                      &       $6.25$                    &           $16.90$                  &               $45.52$               &            $26.43$               &         $18.85$                  &          $\textbf{52.10}$  &           $\textbf{29.59}$     \\

Balanced $|$ Accumulative $|$ 1-hot \& FoodOn    &   $\textbf{10.99}$                    &        $\textbf{20.82}$                      &       $\textbf{14.59}$                    &           $\textbf{18.97}$                  &             $\textbf{51.19}$                 &           $\textbf{29.67}$                &         18.85                    &       $\textbf{52.10}$   &          $\textbf{29.59}$    \\

\end{tabular}
\end{adjustbox}
\vspace{-0.3cm}
\end{table*}

%\paragraph{How Many Examples to Get to Best Performance?}

%Table \ref{tab:tutoring-method} shows the agent's complete learning curve across the training set for 100, 10k, and 49k examples provided.
%
Moreover, we can notice that when the agent is provided with the examples in \textit{Balanced} order and learns with the \textit{Accumulative} method, it reaches around half the task performance with as few as $100$ examples ($0.2\%$ of the dataset), compared to being provided with all the training examples ($49$k).
Additionally, it reaches the complete dataset performance in the first $10$k ($20\%$ of the dataset) examples .
In contrast, the \textit{Baseline} method does not reach its half performance when provided $100$ examples with the \textit{Balanced} method, except when evaluating with Hit@1 ($9.05\%$ instead of $18.93\%$), and it does not reach the complete dataset performance when provided with $10$k examples.
This, again, suggests that the \textit{Accumulative} method can generalize faster, and can treat the $80\%$ of the dataset as redundant when tutored with the \textit{Balanced} policy.
When the examples are given in random order, no learning method achieves complete dataset performance before using all the training examples.

These findings indicate that the order in which the human provides the examples has a great impact on how quickly the agent can pick up the task.
\textit{A tutoring policy that balances repeating dominant examples and providing diverse examples can double or triple the speed at which the agent learns how to perform the task for our use case.}
\vspace{-0.4cm}
%%%%%%%%%%%%%%%%%%%%%%%%%%%%%%%%%%%%%%%%%%%%%%%%%%%%%%%%%%%%%%%%%%%%%%%%
\section{Conclusions and Future Work}
\label{sec:conclusion}
\vspace{-0.2cm}

In this work, we proposed an intuitive human-agent teaching architecture that allows teaching an agent how to perform a task from a few examples. 
Under this perspective, a human can serve as a teacher, helping the agent to learn new concepts efficiently. 
Specifically, the human can provide a small and incremental number of significative examples to the agent, allowing the agent to reach new knowledge in a faster way. 
On the other hand, the agent can leverage its computational capabilities to learn from the examples provided by the human.
To apply our architecture, we propose a use case where the agent has to learn how to mimic the way in which a human substitutes ingredients in recipes, using as few provided examples as possible.
We replicate human input with the Recipe1MSubs dataset and experimentally test our approach.
We use this dataset as it is the first dataset consisting of real ingredient substitution suggestions from human users.
We have three key findings from our experiments, one for each component (research question).
First, symbolic ingredient knowledge, i.e., FoodOn, allows for better generalization compared to using implicit ingredient semantics captured by sub-symbolic embeddings, i.e., FoodBert and FlavorGraph.
Second, we put forward a learning method that allows for efficient and incremental learning.
Third, we measure the effect of potential learning efficiency of the order in which the examples are provided to the agent and suggest a tutoring policy that assists the agent in achieving half its performance after only 100 examples.

% % 
% \begin{inlinelist}
%     \item \ref{rq1}, symbolic ingredient knowledge, i.e., FoodOn, allows for better generalization compared to using implicit ingredient semantics captured by sub-symbolic embeddings, i.e., FoodBert and FlavorGraph.
%     \item \ref{rq2}, we put forward a few-shot learning method that allows for efficient and incremental learning.
%     \item \ref{rq3}, we measure the effect of potential learning efficiency of the order in which the examples are provided to the agent and suggest a tutoring policy that assists the agent in achieving half its performance after only 100 examples.
% \end{inlinelist} 

Our study can be extended in several ways.
First, the proposed methods should be tested using data from individual users.
Second, a user survey should also evaluate their efficiency and intuitiveness, in terms of total effort required from users to teach the agent.
%
%Third, methods that allow for richer representations of the recipes and the substitution queries, e.g., \cite{ShiraiK22}, should be investigated to take into account the cooking actions.
% 
Third, the agent should use an AL method, guiding the user to provide examples according to its current strengths and weaknesses in the learning process.
% 
%Fifth, the Recipe1MSubs dataset should be extended to have more labels, allowing for better comparison among learning methods.
% 
Last but not least, the overall intuitive human-agent interaction architecture should be tested to a different use case, e.g., in an educational setting where the agent needs to understand the hobbies and interests of the students in order to accordingly tailor exercises to keep the students interested, or to propose media to which students react the best.

\vspace{-0.1cm}

\begin{credits}
\subsubsection*{\ackname}

This work was supported by ``MUHAI - Meaning and Understanding in Human-centric Artificial Intelligence'' project, funded by the European Union's Horizon 2020 research and innovation program under grant agreement No 951846.
\begin{minipage}{0.05\textwidth}
  \includegraphics[trim={0cm 0cm 0cm 0cm},width=\textwidth]{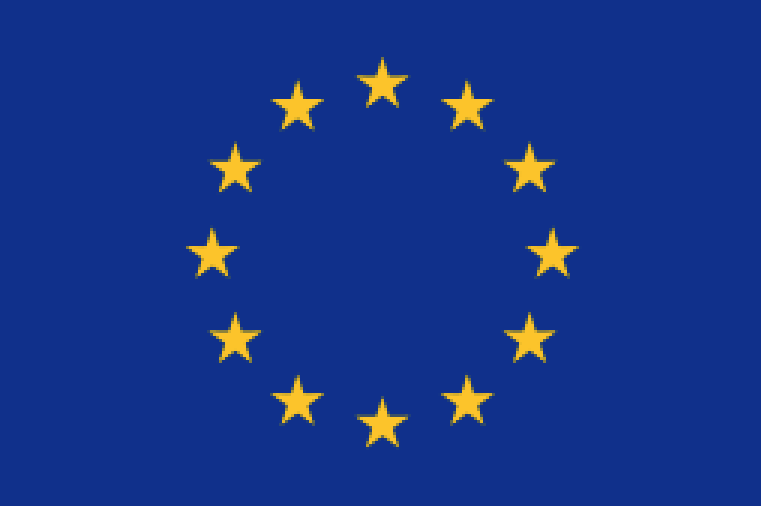}
\end{minipage}%

\noindent

\end{credits}

\newpage

\bibliographystyle{splncs04} 
\bibliography{bibliography}
\end{document}